\documentclass[11pt,a4paper]{article}
\usepackage[hyperref]{emnlp2020}
\usepackage{times}
\usepackage{latexsym}
\usepackage{graphicx}
\usepackage{subcaption}
\usepackage{svg}
\usepackage{amsmath}
\usepackage{lipsum}
\usepackage{enumitem}

\usepackage{microtype}

\aclfinalcopy 

\title{Not Low-Resource Anymore: Aligner Ensembling, Batch Filtering, and New Datasets for Bengali-English Machine Translation}

\author{Tahmid Hasan\thanks{~ These authors contributed equally to this work.} , Abhik Bhattacharjee\footnotemark[1] , Kazi Samin, Masum Hasan, Madhusudan Basak,\\
{\bf M. Sohel Rahman and Rifat Shahriyar}\\
  Bangladesh University of Engineering and Technology (BUET)\\
  \texttt{\{tahmidhasan, madhusudan, msrahman, rifat\}@cse.buet.ac.bd,}\\
  \texttt{\{abhik, samin, masum\}@ra.cse.buet.ac.bd}}
  
\date{}
\begin{document}
\maketitle
\begin{abstract}
Despite being the seventh most widely spoken language in the world, Bengali has received much less attention in machine translation literature due to being low in resources. Most publicly available parallel corpora for Bengali are not large enough; and have rather poor quality, mostly because of incorrect sentence alignments resulting from erroneous sentence segmentation, and also because of a high volume of noise present in them. In this work, we build a customized sentence segmenter for Bengali and propose two novel methods for parallel corpus creation on low-resource setups: aligner ensembling and batch filtering. With the segmenter and the two methods combined, we compile a high-quality Bengali-English parallel corpus comprising of 2.75 million sentence pairs, more than 2 million of which were not available before. Training on neural models, we achieve an improvement of more than 9 BLEU score over previous approaches to Bengali-English machine translation. We also evaluate on a new test set of 1000 pairs made with extensive quality control. We release the segmenter, parallel corpus, and the evaluation set, thus elevating Bengali from its low-resource status. To the best of our knowledge, this is the first ever large scale study on Bengali-English machine translation. We believe our study will pave the way for future research on Bengali-English machine translation as well as other low-resource languages. Our data and code are available at \url{https://github.com/csebuetnlp/banglanmt}.
\end{abstract} 
\section{Introduction}
Recent advances in deep learning \citep{bahdanau2014neural, wu2016google, vaswani2017attention} have aided in the development of neural machine translation (NMT) models to achieve state-of-the-art results in several language pairs. But a large number of high-quality sentence pairs must be fed into these models to train them effectively \citep{koehn2017six}; and in fact lack of such a corpus affects the performance thereof severely. Although there have been efforts to improve machine translation in low-resource contexts, particularly using, for example, comparable corpora \citep{irvine2013combining}, small parallel corpora \citep{gu2018universal} or zero-shot multilingual translation \citep{johnson2017google}, such languages are yet to achieve noteworthy results \citep{koehn2019findings} compared to high-resource ones. Unfortunately, Bengali, the seventh (fifth) most widely spoken language in the world by the number of (native\footnote{\url{https://w.wiki/Psq}}) speakers,\footnote{\url{https://w.wiki/Pss}} has still remained a low-resource language. As of now, only a few parallel corpora for Bengali language are publicly available \citep{tiedemann2012parallel} and those too suffer from poor sentence segmentation, resulting in poor alignments. They also contain much noise, which, in turn, hurts translation quality \citep{khayrallah2018impact}. No previous work on Bengali-English machine translation addresses any of these issues.

With the above backdrop, in this work, we develop a customized sentence segmenter for Bengali language while keeping uniformity with the English side segmentation. We experimentally show that better sentence segmentation that maintains homogeneity on both sides results in better alignments. We further empirically show that the choice of sentence aligner plays a significant role in the quantity of parallel sentences extracted from document pairs. In particular, we study three aligners and show that combining their results, which we name `Aligner Ensembling', increases recall. We introduce `Batch Filtering', a fast and effective method for filtering out incorrect alignments. Using our new segmenter, aligner ensemble, and batch filter, we collect a total of 2.75 million high-quality parallel sentences from a wide variety of domains, more than 2 million of which were not previously available. Training our corpus on NMT models, we outperform previous approaches to Bengali-English machine translation by more than 9 BLEU \citep{papineni2002bleu} points and also show competitive performance with automatic translators. We also prepare a new test corpus containing 1000 pairs made with extensive manual and automated quality checks. Furthermore, we perform an ablation study to validate the soundness of our design choices. 

We release all our tools, datasets, and models for public use. To the best of our knowledge, this is the first ever large scale study on machine translation for Bengali-English pair. We believe that the insights brought to light through our work may give new life to Bengali-English MT that suffered so far for being low in resources. We also believe that our findings will also help design more efficient methods for other low-resource languages.

\section{Sentence Segmentation}
Proper sentence segmentation is an essential pre-requisite for sentence aligners to produce coherent alignments. However, segmenting a text into sentences is not a trivial task, since the end-of-sentence punctuation marks are ambiguous. For example, in English, the end-of-sentence period, abbreviations, ellipsis, decimal point, etc. use the same symbol (.). Since either side of a document pair can contain Bengali/English/foreign text, we need a sentence segmenter to produce consistent segmentation in a language-independent manner.

\begin{figure}[h]
  \centering
  \includegraphics[width=0.5\textwidth]{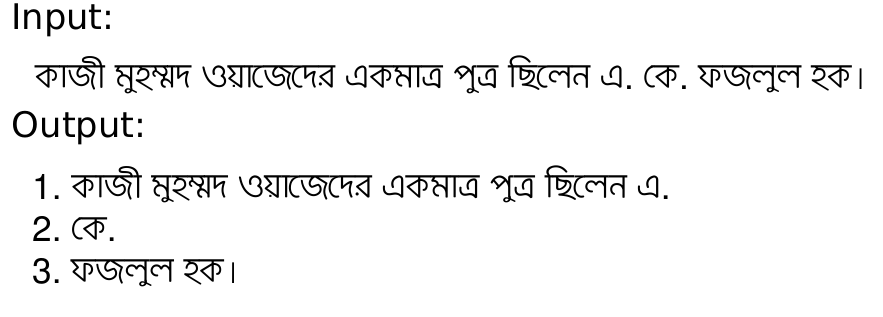} 
  \caption{Erroneous sentence segmentation by Polyglot}
  \label{polyglot}
\end{figure}

Available libraries supporting both Bengali and English segmentation, e.g., Polyglot \citep{al2013polyglot}, do not work particularly well for Bengali sentences with abbreviations, which is common in many domains. For instance, Polyglot inaccurately splits the input sentence in Figure \ref{polyglot} into three segments,
whereas the English side can successfully detect the non-breaking tokens. Not only does this corrupt the first alignment, but also causes the two broken pieces to be aligned with other sentences, creating a chain of incorrect alignments.

SegTok,\footnote{\url{https://github.com/fnl/segtok}} a rule-based segmentation library, does an excellent job of segmenting English texts. SegTok uses regular expressions to handle many complex cases, e.g., technical texts, URLs, abbreviations. We extended SegTok's code to have the same functionality for Bengali texts by adding new rules (e.g., quotations, parentheses, bullet points) and abbreviations identified through analyzing both Bengali and English side of our corpus, side-by-side enhancing SegTok's English segmentation correctness as well. Our segmenter can now address the issues like the example mentioned and provide consistent outputs in a language-agnostic manner.

We compared the performance of our segmenter on different aligners against Polyglot. We found that despite the number of aligned pairs decreased by 1.37\%, the total number of words on both sides increased by 5.39\%, making the resulting parallel corpus richer in content than before. This also bolsters our hypothesis that Polyglot creates unnecessary sentence fragmentation. 
\section{Aligner Selection and Ensembling}
\label{sec:aligner}
\subsection{Aligner Descriptions}
\label{sec:alignerd}
Most available resources for building parallel corpora come in the form of parallel documents which are exact or near-exact translations of one another. Sentence aligners are used to extract parallel sentences from them, which are then used as training examples for MT models. \citet{abdul2012extrinsic} conducted a comparative evaluation of five aligners and showed that the choice of aligner had considerable performance gain by the models trained on the resultant bitexts. They identified three aligners with superior performance: Hunalign \citep{varga2007parallel}, Gargantua \citep{braune2010improved}, and Bleualign \citep{sennrich2010mt}.

\begin{figure*}[h]
\begin{subfigure}{0.33\textwidth}
  \centering
  \includegraphics[width=\textwidth]{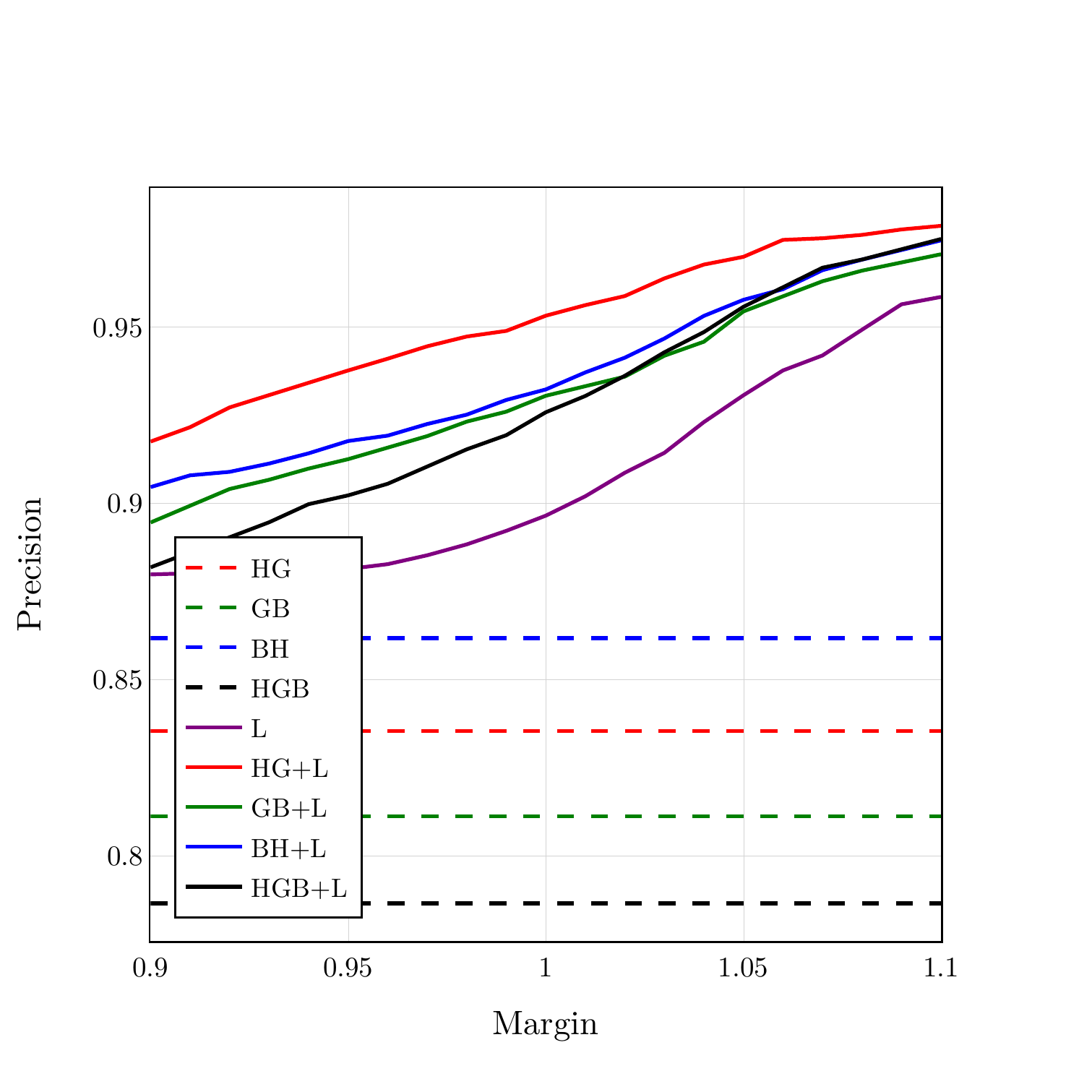}  
  \caption{Precision vs. Margin}
  \label{fig:pvm}
\end{subfigure}
\begin{subfigure}{0.33\textwidth}
  \centering
  \includegraphics[width=\textwidth]{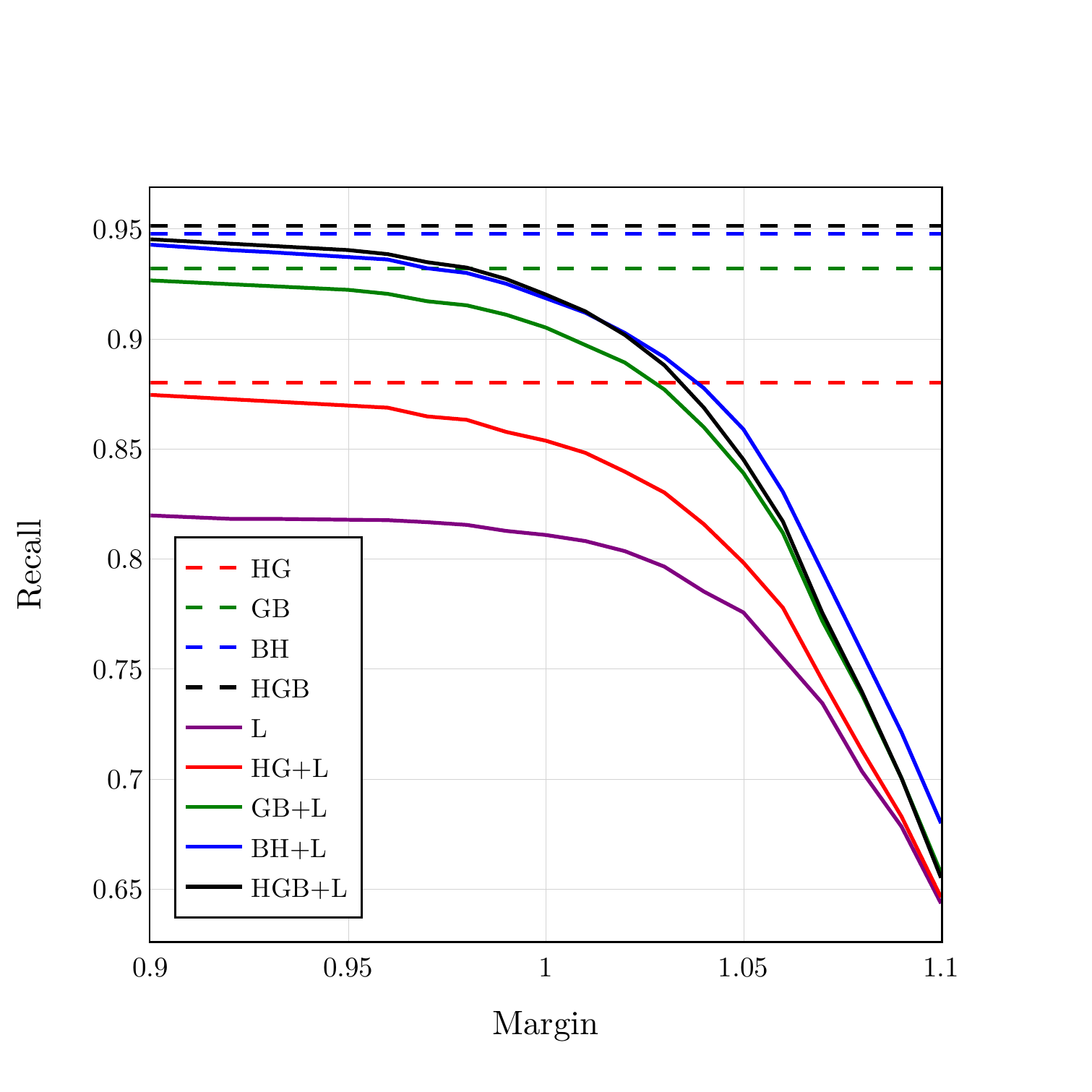}  
  \caption{Recall vs. Margin}
  \label{fig:rvm}
\end{subfigure}
\begin{subfigure}{0.33\textwidth}
  \centering
  \includegraphics[width=\textwidth]{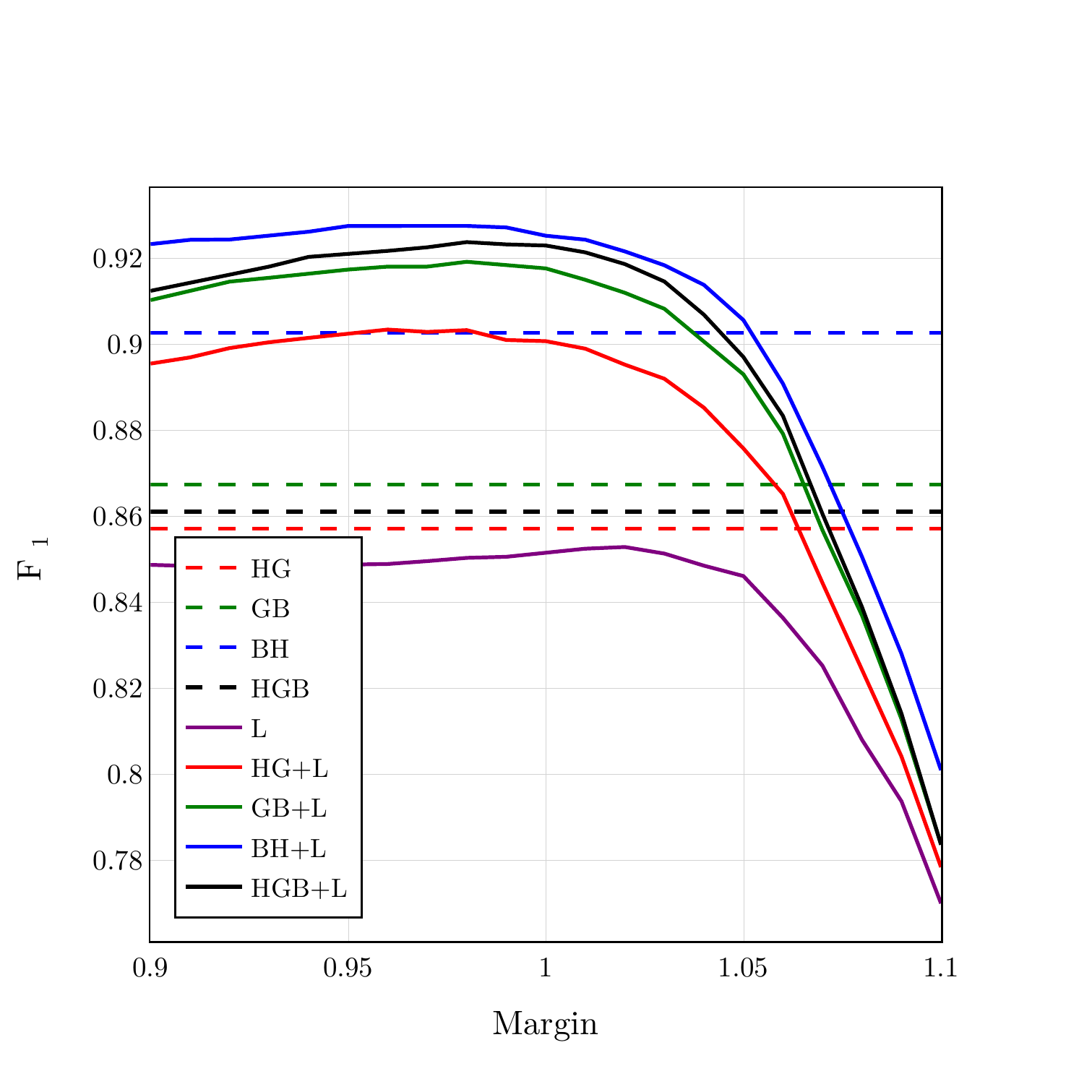}  
  \caption{$F_1$ Score vs. Margin}
  \label{fig:fvm}
\end{subfigure}
\caption{Performance metrics of ensembles with filtering}
\label{fig:margin}
\end{figure*}

However, their results showed performance only in terms of BLEU score, with no indication of any explicit comparison metric between the aligners (e.g., precision, recall). As such, to make an intrinsic evaluation, we sampled 50 documents from four of our sources (detailed in section \ref{sec:docdata}) with their sentence counts on either side ranging from 20 to 150. We aligned sentences from these documents manually (i.e., the gold alignment) and removed duplicates, which resulted in 3,383 unique sentence pairs. We then aligned the documents again with the three aligners using our custom segmenter. Table \ref{tab:hgb} shows performance metrics of the aligners.

\subsection{Aligner Ensembling and Filtering}
\label{aligneref}
From the results in Table \ref{tab:hgb}, it might seem that Hunalign should be the most ideal aligner choice. But upon closer inspection, we found that each aligner was able to correctly align some pairs that the other two had failed to do. Since we had started from a low-resource setup, it would be in our best interest if we could combine the data extracted by all aligners. As such, we `ensembled' the results of the aligners as follows. For each combination of the aligners (4 combinations in total; see Table \ref{tab:ensemble}), we took the union of sentence pairs extracted by each constituent aligner of the said combination for each document. The performance of the aligner ensembles is shown in Table \ref{tab:ensemble}. We concatenated the first letters of the constituent aligners to name each ensemble (e.g., HGB refers to the combination of all three of them).

\begin{table}[h]
\centering
\begin{tabular}{l}
\hline
\textbf{Aligner}\\
\hline
Hunalign\\
Gargantua\\
Bleualign\\
\hline
\end{tabular}
\begin{tabular}{ccc}
\hline
\textbf{Precision} & \textbf{Recall} & \textbf{F$_1$}\\
\hline
\textbf{93.21} & 85.82 & \textbf{89.37} \\ 
84.79 & 69.32 & 76.28\\ 
89.41 & \textbf{87.35} & 88.37\\ 
\hline
\end{tabular}
\caption{Performance metrics of aligners}\label{tab:hgb}
\vspace{1em}
\begin{tabular}{l}
\hline
\textbf{Ensemble}\\
\hline
HG\\
GB\\
BH\\
HGB\\
\hline
\end{tabular}
\begin{tabular}{ccc}
\hline
\textbf{Precision} & \textbf{Recall} & \textbf{F$_1$}\\
\hline
83.52 & 88.00 & 85.70\\ 
81.11 & 93.20 & 86.73\\ 
\textbf{86.16} & 94.76 & \textbf{90.26}\\
78.64 & \textbf{95.13} & 86.10\\
\hline
\end{tabular}
\caption{Performance metrics of ensembles}\label{tab:ensemble}
\vspace{1em}
\begin{tabular}{l}
\hline
\textbf{Ensemble}\\
\hline
L(1.02)\\
HG+L(0.96)\\
GB+L(0.98)\\
BH+L(0.96)\\
HGB+L(0.98)\\
\hline
\end{tabular}
\begin{tabular}{ccc}
\hline
\textbf{Precision} & \textbf{Recall} & \textbf{F$_1$}\\
\hline
90.86 & 80.34 & 85.28\\
\textbf{94.09} & 86.86 & 90.33\\ 
92.31 & 91.52 & 91.91\\ 
91.91 & \textbf{93.60} & \textbf{92.75}\\
91.52 & 93.23 & 92.37\\
\hline
\end{tabular}
\caption{Performance metrics of filtered ensembles}\label{tab:fensemble}
\end{table}

Table \ref{tab:ensemble} shows that BH achieved the best $F_1$ score among all ensembles, even 0.89\% above the best single aligner Hunalign. Ensembling increased the recall of BH by 8.94\% compared to Hunalign, but also hurt precision severely (by 7.05\%), due to the accumulation of incorrect alignments made by each constituent aligner. To mitigate this effect, we used the LASER\footnote{\url{https://github.com/facebookresearch/LASER}} toolkit to filter out incorrect alignments. LASER, a cross-lingual sentence representation model, uses similarity scores between the embeddings of candidate sentences to perform as both aligner \citep{schwenk2019wikimatrix} and filter \citep{chaudhary2019low}. We used LASER as a filter on top of the ensembles, varied the similarity margin \citep{artetxe2018margin} between 0.90 to 1.10 with 0.01 increment, and plotted the performance metrics in Figure \ref{fig:margin}. We also reported the performance of LASER as a standalone aligner (referred to as L in the figure; +L indicates the application of LASER as a filter). The dashed lines indicate ensemble performance without the filter.

As Figure \ref{fig:pvm} indicates, ensembles achieve significant gain on precision with the addition of the LASER filter. While recall (Figure \ref{fig:rvm}) doesn't face a significant decline at first, it starts to take a deep plunge when margin exceeds 1.00. We balanced between the two by considering the $F_1$ score (Figure \ref{fig:fvm}). Table \ref{tab:fensemble} shows the performance metrics of LASER and all filtered ensembles for which their respective $F_1$ score is maximized.

Table \ref{tab:fensemble} shows that despite being a good filter, LASER as an aligner does not show considerable performance compared to filtered ensembles. The best $F_1$ score is achieved by the BH ensemble with its margin set to 0.96. Its precision increased by 5.75\% while trailing a mere 1.16\% in recall behind its non-filtered counterpart. Compared to single Hunalign, its recall had a 7.78\% gain, while lagging in precision by only 1.30\%, with an overall $F_1$ score increase of 3.38\%. Thus, in all future experiments, we used BH+L(0.96) as our default aligner with the mentioned filter margin.

\section{Training Data and Batch Filtering}
\label{tdata}

\begin{table*}[h]
\centering
\begin{tabular}{l}
\hline
\textbf{Source}\\
\hline
OpenSubs\\
TED\\
SUPara\\
Tatoeba\\
Tanzil\\
AMARA\\
SIPC\\
Glosbe\\
MediaWiki\\
Gnome\\
KDE\\
Ubuntu\\
Globalvoices\\
JW\\
Banglapedia\\
Books\\
Laws\\
HRW\\
Dictionary\\
Wiki Sections\\
Miscellaneous\\
\hline
\textbf{Total}\\
\hline
\end{tabular}
\begin{tabular}{rrrrr}
\hline
\textbf{\#Pairs} & \textbf{\#Tokens(Bn)} & \textbf{\#Tokens(En)} & \textbf{\#Toks/Sent(Bn)} & \textbf{\#Toks/Sent(En)}\\
\hline
365,837 & 2,454,007 & 2,902,085 & 6.71 & 7.93\\
15,382 & 173,149 & 195,007 & 11.26 & 12.68\\
69,533 & 811,483 & 996,034 & 11.67 & 14.32\\
9,293 & 50,676 & 57,266 & 5.45 & 6.16\\
5,908 & 149,933 & 164,426 & 25.38 & 27.83\\
1,166 & 63,447 & 47,704 & 54.41 & 40.91\\
19,561 & 240,070 & 311,816 & 12.27 & 15.94\\
81,699 & 1,531,136 & 1,728,394 & 18.74 & 21.16\\
45,998 & 3,769,963 & 4,205,913 & 81.96 & 91.44\\
102,078 & 725,297 & 669,659 & 7.11 & 6.56\\
16,992 & 122,265 & 115,908 & 7.20 & 6.82\\
5,251 & 22,727 & 22,616 & 4.33 & 4.29\\
235,106 & 4,162,896 & 4,713,335 & 17.70 & 20.04\\
546,766 & 9,339,929 & 10,215,160 & 17.08 & 18.68\\
264,043 & 3,695,930 & 4,643,818 & 14.00 & 17.59\\
99,174 & 1,393,095 & 1,787,694 & 14.05 & 18.03\\
28,218 & 644,384 & 801,092 & 22.84 & 28.39\\
2,586 & 55,469 & 65,103 & 21.44 & 25.17\\
483,174 & 700,870 & 674,285 & 1.45 & 1.40\\
350,663 & 5,199,814 & 6,397,595 & 14.83 & 18.24\\
2,877 & 21,427 & 24,813 & 7.45 & 8.62\\
\hline
\textbf{2,751,315} & \textbf{35,327,967} & \textbf{40,739,723} & \textbf{12.84} & \textbf{14.81}\\
\hline
\end{tabular}
\label{tab:dataset}
\caption{Summary of the training corpus.}\label{tab:data}
\end{table*}

We categorize our training data into two sections: (1) Sentence-aligned corpora and (2) Document-aligned corpora.

\subsection{Sentence-aligned Corpora}\label{sec:sent}
We used the corpora mentioned below which are aligned by sentences:

\begin{enumerate}[label={}, leftmargin=*]
 \setlength{\itemsep}{1pt}
 \item \textbf{Open Subtitles 2018} corpus \citep{lison2019open} from OPUS\footnote{\url{opus.nlpl.eu}} \citep{tiedemann2012parallel}
 \item \textbf{TED} corpus \citep{cettolo2012wit3}
 \item \textbf{SUPara} corpus \citep{al2012supara}
 \item \textbf{Tatoeba} corpus from \url{tatoeba.org}
 \item \textbf{Tanzil} corpus from the Tanzil project\footnote{\url{tanzil.net/docs/tanzil_project}}
 \item \textbf{AMARA} corpus \citep{abdelali2014amara}
 \item \textbf{SIPC} corpus \citep{post2012constructing}
 \item \textbf{Glosbe}\footnote{\url{https://glosbe.com/}} online dictionary example sentences
 \item \textbf{MediaWiki Content Translations}\footnote{\url{https://w.wiki/RZn}}
 \item \textbf{Gnome, KDE, Ubuntu} localization files
 \item \textbf{Dictionary} entries from \url{bdword.com}
 \item \textbf{Miscellaneous} examples from \url{english-bangla.com} and \url{onubadokderadda.com}
\end{enumerate}

\subsection{Document-aligned Corpora}\label{sec:docdata}
The corpora below have document-level links from where we sentence-aligned them:

\begin{enumerate}[label={}, leftmargin=*]
 \item \textbf{Globalvoices:} Global Voices\footnote{\url{https://globalvoices.org/}} publishes and translates articles on trending issues and stories from press, social media, blogs in more than 50 languages. Although OPUS provides sentence-aligned corpus from Global Voices, we re-extracted sentences using our segmenter and filtered ensemble, resulting in a larger amount of pairs compared to OPUS.
 
 \item \textbf{JW:} \citet{agic2019jw300} introduced JW300, a parallel corpus of over 300 languages crawled from \url{jw.org}, which also includes Bengali-English. They used Polyglot \citep{al2013polyglot} for sentence segmentation and Yasa \citep{lamraoui2013yet} for sentence alignment. We randomly sampled 100 sentences from their Bengali-English corpus and found only 23 alignments to be correct. So we crawled the website using their provided instructions and aligned using our segmenter and filtered ensemble. This yielded more than twice the data than theirs. 

 \item \textbf{Banglapedia:} ``Banglapedia: the National Encyclopedia of Bangladesh" is the first Bangladeshi encyclopedia. Its online version\footnote{\url{https://www.banglapedia.org/}} contains over 5,700 articles in both Bengali and English. We crawled the website to extract the article pairs and aligned sentences with our segmenter and filtered ensemble.
 
 \item \textbf{Bengali Translation of Books:} We collected translations of more than 100 books available on the Internet with their genres ranging from classic literature to motivational speeches and aligned them using our segmenter and filtered ensemble.
 
 \item \textbf{Bangladesh Law Documents:} The Legislative and Parliamentary Affairs Division of Bangladesh makes all laws available on their website.\footnote{\url{bdlaws.minlaw.gov.bd}} Some older laws are also available under the ``Heidelberg Bangladesh Law Translation Project".\footnote{\url{https://www.sai.uni-heidelberg.de/workgroups/bdlaw/}} Segmenting the laws was not feasible with the aligners in section \ref{sec:alignerd} as most lines were bullet points terminating in semicolons, and treating semicolons as terminals broke down valid sentences. Thus, we made a regex-based segmenter and aligner for these documents. Since most laws were exact translations with an equal number of bullet points under each section, the deterministic aligner yielded good alignment results.
 
 \item \textbf{HRW:} Human Rights Watch\footnote{\url{https://www.hrw.org/}} investigates and reports on abuses happening in all corners of the world on their website. We crawled the Bengali-English article pairs and aligned them using our segmenter and filtered ensemble.
 
 \item \textbf{Wiki Sections:} Wikipedia is the largest multilingual resource available on the Internet. But most article pairs are not exact or near-exact translations of one another. However, such a large source of parallel texts cannot be discarded altogether. Wikimatrix \citep{schwenk2019wikimatrix} extracted bitexts from Wikipedia for 1620 language pairs, including Bengali-English. But we found them to have issues like foreign texts, incorrect sentence segmentations and alignments etc. As such, we resorted to the original source and only aligned from sections having high similarity. We translated the Bengali articles into English using an NMT model trained on the rest of our data and compared each section of an article against the sections of its English counterpart. We used SacreBLEU \citep{post2018call} score as the similarity metric and only picked sections with score above 20. We then used our filtered ensemble on the resulting matches. 
\end{enumerate}

\subsection{Batch Filtering}
LASER uses cosine similarity between candidate sentences as the similarity metric and calculates margin by normalizing over the nearest neighbors of the candidates. \citet{schwenk2019wikimatrix} suggested using a global space, i.e., the complete corpus for neighbor search while aligning, albeit without any indication of what neighborhood to use for filtering. In section \ref{aligneref}, we used local neighborhood on document level and found satisfactory results. So we tested it with a single aligner, Hunalign,\footnote{The optimal margin was found to be 0.95 for Hunalign.} on three large document sources, namely, Globalvoices (GV), JW, and Banglapedia (BP). But the local approach took over a day to filter from about 25k document pairs, the main bottleneck being the loading time for each document. Even with several optimizations, running time did not improve much. The global approach suffered from another issue: memory usage. The datasets were too large to be fit into GPU as a whole.\footnote{We used an RTX 2070 GPU with 8GB VRAM for these experiments.} Thus, we shifted the neighbor search to CPU, but that again took more than a day to complete. Also, the percentage of filtered pairs was quite higher than the local neighborhood approach, raising the issue of data scarcity again. So, we sought the following middle-ground between global and local approach: for each source, we merged all alignments into a single file, shuffled all pairs, split the file into 1k size batches, and then applied LASER locally on each batch, reducing running time to less than two hours.

\begin{table}[h]
\centering
\begin{tabular}{l}
\hline
\textbf{Source}\\
\hline
GV\\
JW\\
BP\\
\hline
\end{tabular}
\begin{tabular}{ccc}
\hline
\textbf{Document} & \textbf{1k Batch} & \textbf{Global}\\
\hline
\textbf{4.05} & 4.60 & 8.03\\ 
\textbf{6.22} & 7.06 & 13.28\\ 
\textbf{13.01} & 14.96 & 25.65\\ 
\hline
\end{tabular}
\caption{Filtered pairs (\%) for different neighborhoods}\label{tab:yield}
\end{table}

In Table \ref{tab:yield}, we show the percentage of filtered out pairs from the sources for each neighborhood choice. The global approach lost about twice the data compared to the other two. The 1k batch neighborhood achieved comparable performance with respect to the more fine-grained document-level neighborhood while improving running time more than ten-folds. Upon further inspection, we found that more than 98.5\% pairs from the document-level filter were present in the batched approach. So, in subsequent experiments, we used `Batch Filtering' as standard. In addition to the document-aligned sources, we also used batch filtering on each sentence-aligned corpus in section \ref{sec:sent} to remove noise from them. Table \ref{tab:data} summarizes our training corpus after the filtering.

\section{Evaluation Data}

A major challenge for low-resource languages is the unavailability of reliable evaluation benchmarks that are publicly available. After exhaustive searching, we found two decent test sets and developed one ourselves. They are mentioned below:
\begin{enumerate}[label={}, leftmargin=*]
 \item \textbf{SIPC:} \citet{post2012constructing} used crowdsourcing to build a collection of parallel corpora between English and six Indian languages, including Bengali. Although they are not translated by experts and have issues for many sentences (e.g., all capital letters on English side, erroneous translations, punctuation incoherence between Bn and En side, presence of foreign texts), they provide four English translations for each Bengali sentence, making it an ideal test-bed for evaluation using multiple references. We only evaluated the performance of Bn$\rightarrow$En for this test set.
 \item \textbf{SUPara-benchmark} \citep{czes-gs42-18}: Despite having many spelling errors, incorrect translations, too short (less than 50 characters) and too long sentences (more than 500 characters), due to its balanced nature having sentences from a variety of domains, we used it for our evaluation.
 \item \textbf{RisingNews:} Since the two test sets mentioned above suffer from many issues, we created our own test set. Risingbd,\footnote{\url{https://www.risingbd.com/}} an online news portal in Bangladesh, publishes professional English translations for many of their articles. We collected about 200 such article pairs and had them aligned by an expert. We had them post-edited by another expert. We then removed, through automatic filtering, pairs that had (1) less than 50 or more than 250 characters on either side, (2) more than 33\% transliterations or (3) more than 50\% or more than 5 OOV words \citep{guzman2019two}. This resulted in 600 validation and 1000 test pairs; we named this test set ``\textbf{RisingNews}".
\end{enumerate}
\section{Experiments and Results}

\subsection{Pre-processing}
Before feeding into the training pipeline, we performed the following pre-processing sequentially:
\begin{enumerate}[leftmargin=*]
\setlength{\itemsep}{1pt}
\item We normalized punctuations and characters that have multiple unicode representations to reduce data sparsity.
\item We removed foreign strings that appear on both sides of a pair, mostly phrases from which both sides of the pair have been translated.
\item We transliterated all dangling English letters and numerals on the Bn side into Bengali, mostly constituting bullet points.
\item Finally, we removed all evaluation pairs from the training data to prevent data leakage.
\end{enumerate}

At this point, a discussion with respect to language classification is in order. It is a standard practice to use a language classifier (e.g., \citealp{joulin2016bag}) to filter out foreign texts. But when we used it, it classified a large number of valid English sentences as non-English, mostly because they contained named entities transliterated from Bengali side. Fearing that this filtering would hurt translation of named entities, we left language classification out altogether. Moreover, most of our sources are bilingual and we explicitly filtered out sentences with foreign characters, so foreign texts would be minimal.

As for the test sets, we performed minimal pre-processing: we applied character and punctuation normalization; and since SIPC had some sentences that were all capital letters, we lowercased those (and those only).

\subsection{Comparison with Previous Results}
We compared our results with \citet{mumin2019shutorjoma},  \citet{hasanneural}, and \citet{mumin2019neural}. The first work used SMT, while the latter two used NMT models. All of them evaluated on the SUPara-benchmark test set. We used the OpenNMT \citep{klein2017opennmt} implementation of big Transformer model \citep{vaswani2017attention} with 32k vocabulary on each side learnt by Unigram Language Model with subword regularization\footnote{l=32, $\alpha$=0.1} \citep{kudo2018subword} and tokenized using SentencePiece \citep{kudo2018sentencepiece}. To maintain consistency with previous results, we used lowercased BLEU \citep{papineni2002bleu} as the evaluation metric. Comparisons are shown in Table \ref{tab:supara}. 

\begin{table}[h]
\centering
\begin{tabular}{l}
\hline
\textbf{Model}\\
\hline
\citet{mumin2019shutorjoma}\\
\citet{hasanneural}\\
\citet{mumin2019neural}\\
Ours\\
\hline
\end{tabular}
\begin{tabular}{cc}
\hline
\textbf{Bn$\rightarrow$En} & \textbf{En$\rightarrow$Bn}\\
\hline
17.43 & 15.27\\ 
19.98 & --\\ 
22.68 & 16.26\\
\textbf{32.10} & \textbf{22.02}\\
\hline
\end{tabular}
\caption{Comparison (BLEU) with previous works on SUPara-benchmark test set (\citealp{hasanneural} did not provide En$\rightarrow$Bn scores)}\label{tab:supara}
\end{table}

Evident from the scores in Table \ref{tab:supara}, we outperformed all works by more than \textbf{9} BLEU points for Bn$\rightarrow$En. Although for En$\rightarrow$Bn the difference in improvement (\textbf{5.5+}) is not that much striking compared to Bn$\rightarrow$EN, it is, nevertheless, commendable on the basis of Bengali being a morphologically rich language.

\subsection{Comparison with Automatic Translators}
We compared our models' SacreBLEU\footnote{BLEU+case.mixed+numrefs.1+smooth.exp+tok.13a +version.1.4.1 (numrefs.4 for SIPC)} \citep{post2018call} scores with Google Translate and Bing Translator, two most widely used publicly available automatic translators. Results are shown in Table \ref{tab:auto}.

\begin{table}[h]
\centering
\begin{tabular}{l}
\hline
\textbf{Model}\\
\textbf{/Translator}\\
\hline
Google\\
Bing\\
Ours\\
\hline
\end{tabular}
\begin{tabular}{ccc}
\hline
\textbf{SUPara} & \textbf{SUPara} & \textbf{SIPC}\\
\textbf{Bn$\rightarrow$En} & \textbf{En$\rightarrow$Bn} & \textbf{Bn$\rightarrow$En}\\
\hline
29.4 & 11.1 & 41.2\\ 
24.4 & 10.7 & 37.2\\
\textbf{30.7} & \textbf{22.0} & \textbf{42.7}\\
\hline
\end{tabular}
\caption{Comparison (SacreBLEU) with automatic translators}\label{tab:auto}
\end{table}

From Table \ref{tab:auto} we can see that our models have superior results on all test sets when compared to Google and Bing.

\subsection{Evaluation on RisingNews}
We performed evaluation on our own test set, \textbf{RisingNews}. We show our models' lowercased detokenized BLEU and mixedcased SacreBLEU scores in Table \ref{tab:newstest}.

\begin{table}[h]
\centering
\begin{tabular}{l}
\hline
\textbf{Metric\iffalse /Translator\fi}\\
\hline
BLEU\\
SacreBLEU\\
\hline
\end{tabular}
\begin{tabular}{cc}
\hline
\textbf{Bn$\rightarrow$En} & \textbf{En$\rightarrow$Bn}\\
\hline
39.04 & 27.73\\
36.1 & 27.7\\
\hline
\end{tabular}
\caption{Evaluation on \textbf{RisingNews} corpus} \label{tab:newstest}
\end{table}

\begin{figure*}[h]
  \centering
  \includegraphics{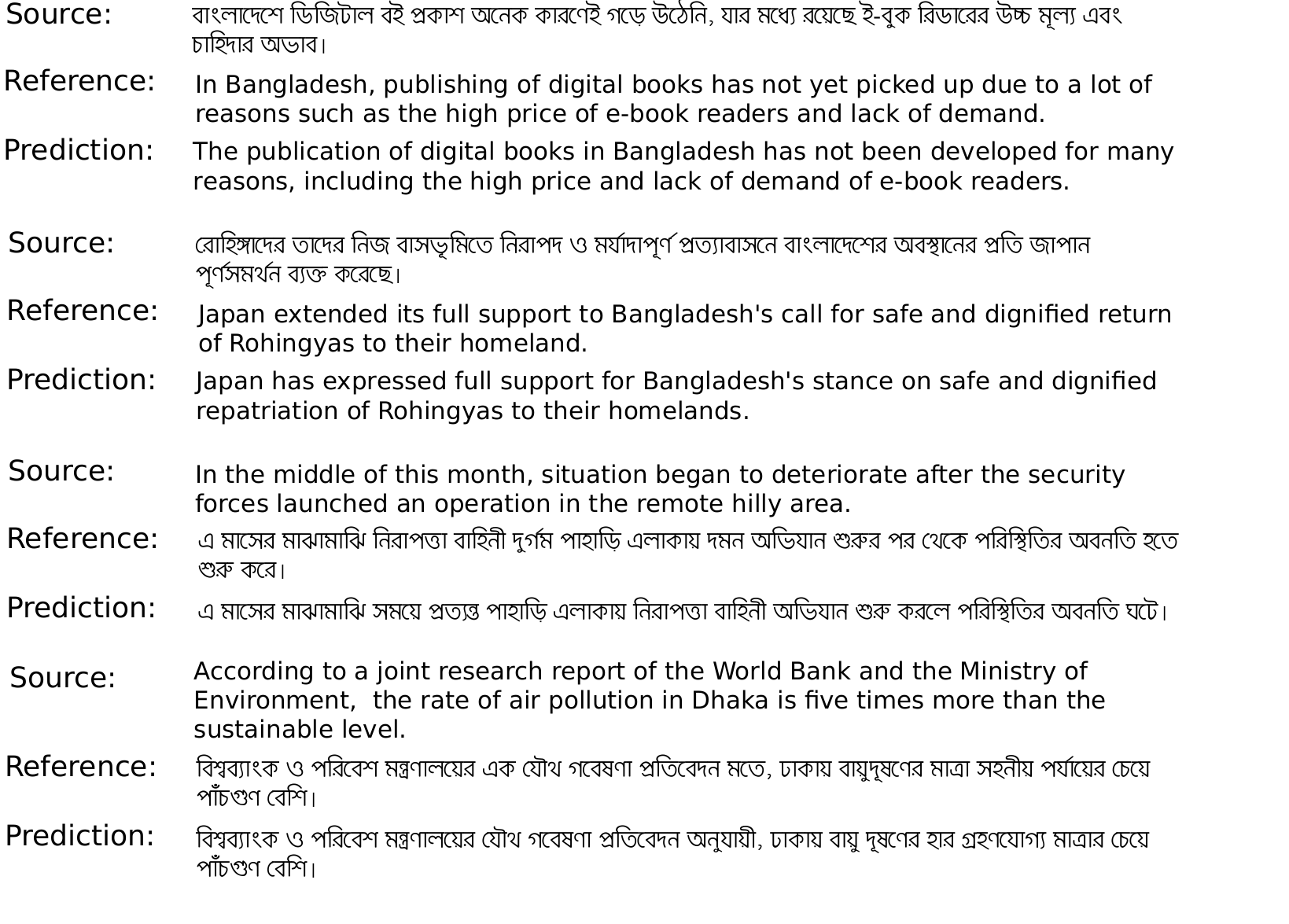}  
  \caption{Sample translations from the RisingNews test set}
  \label{examples}
\end{figure*}

We put great care in creating the test set by performing extensive manual and automatic quality control, and believe it is better in quality than most available evaluation sets for Bengali-English. We also hope that our performance on this test set will act as a baseline for future works on Bengali-English MT. In Figure \ref{examples}, we show some example translations from the RisingNews test set.

\subsection{Comparison with Human Performance}
Remember that SIPC had four reference English translations for each Bengali sentence. We used the final translation as a baseline human translation and used the other three as ground truths (the fourth reference had the best score among all permutations). To make a fair comparison, we evaluated our model's score on the same three references instead of four. Human SacreBLEU score was 32.6, while our model scored 38.0, about \textbf{5.5} points above human judgement.

\subsection{Ablation Study of Filtered Ensembles}
To validate that our choice of ensemble and filter had direct impact on translation scores, we performed an ablation study. We chose four combinations based on their $F_1$ scores from section \ref{sec:aligner}:
\begin{enumerate}[leftmargin=*]
\setlength{\itemsep}{1pt}
\item Best aligner: \textbf{Hunalign}
\item Best aligner with filter: \textbf{Hunalign+L(0.95)}
\item Best ensemble: \textbf{BH}
\item Best ensemble with filter: \textbf{BH+L(0.96)}
\end{enumerate}

To ensure apples to apples comparison, we only used data from the parallel documents, i.e., Globalvoices, JW, Banglapedia, HRW, Books, and Wiki sections. Table \ref{tab:ablation} shows  SacreBLEU scores along with the number of pairs for these combinations. We used the base Transformer model.

\begin{table}[h]
\centering
\begin{tabular}{l}
\hline
\textbf{Aligner}\\
\textbf{/Ensemble}\\
\hline
Hunalign\\
H+L(.95)\\
BH\\
BH+L(.96)\\
\hline
\end{tabular}
\begin{tabular}{ccc}
\hline
\textbf{\#Pairs} & \textbf{SUPara} & \textbf{SIPC}\\
\textbf{(million)} & \textbf{Bn$\rightarrow$En} & \textbf{Bn$\rightarrow$En}\\
\hline
1.35 & 20.5 & 33.2\\ 
1.20 & 21.0 & 33.9\\ 
\textbf{1.64} & 21.0 & 34.0\\ 
1.44 & \textbf{22.1} & \textbf{35.7}\\ 
\hline
\end{tabular}
\caption{SacreBLEU scores for ablation study}\label{tab:ablation}
\end{table}

\begin{figure}[h]
  \centering
  \includegraphics[width=.5\textwidth]{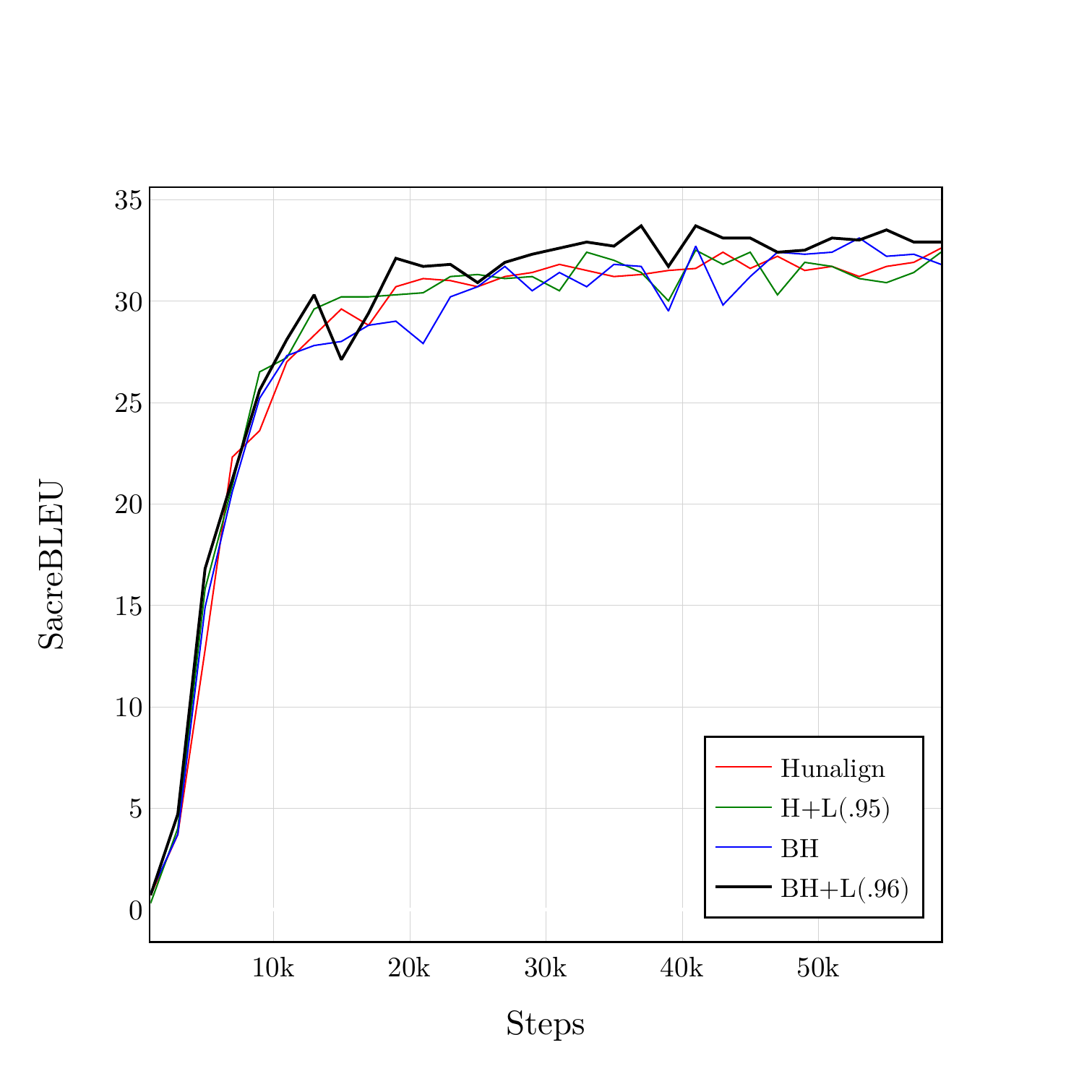}  
  \caption{SacreBLEU vs Steps on SIPCdev set}\label{fig:ablation}
\end{figure}

BH+L(.96) performed better than others by a noticeable margin, and the single Hunalign performed the poorest. While only having 73\% pairs compared to BH, H+L(.95) stood almost on par. Despite the superiority in data count, BH could not perform well enough due to the accumulation of incorrect alignments from its constituent aligners. A clearer picture can be visualized through Figure \ref{fig:ablation}. BH+L(.96) mitigated both data shortage and incorrect alignments and formed a clear envelope over the other three, giving clear evidence that the filter and the ensemble complemented one another.
\vspace{-2ex}

\section{Related Works}
The first initiative towards machine translation for Bengali dates back to the 90s. \citet{sinha1995anglabharti} developed ANGLABHARTI, a rule-based translation system from English to multiple Indian languages, including Bengali. \citet{asaduzzaman2003morphological, dasgupta2004optimal} conducted extensive syntactic analyses to write rules for constructing Bengali parse trees and designed algorithms to transfer between Bengali and English parse trees.

Subsequently, \citet{saha2005semantics} reported an example-based machine translation approach for translating news headlines using a knowledge base. \citet{naskar2006phrasal} described a hybrid between rule-based and example-based translation approaches; here terminals would end at phrases that would then be looked up in the knowledge base.

The improved translation quality of phrase-based statistical machine translation (SMT) \citep{koehn2003statistical} and the wide availability of toolkits thereof \citep{koehn2007moses} created an increased interest in SMT for Bangali-English. As SMT was more data-driven, specialized techniques were integrated to account for the low amount of parallel data for Bengali-English. Among many, \citet{roy2009semi} proposed several semi-supervised techniques; \citet{haffari2009active} used active learning to improve SMT; \citet{islam2010english} used an additional transliteration module to handle OOV words; \citet{banerjee2018multilingual} introduced multilingual SMT for Indic languages, including Bengali.

Although NMT is currently being hailed as the state-of-the-art, very few works have been done on NMT for the Bengali-English pair. \citet{dandapat2018training} trained a deployable general domain NMT model for Bengali-English using sentences aligned from comparable corpora. They combated the inadequacy of training examples by data augmentation using back-translation \citep{sennrich2015improving}. \citet{hasanneural, mumin2019neural} also showed with limited parallel data available on the web that NMT provided improved translation for Bengali-English pair.
\section{Conclusion and Future Works}
In this work, we developed a custom sentence segmenter for Bengali, showed that aligner ensembling with batch filtering provides better performance than single sentence aligners, collected a total of 2.75 million high-quality parallel sentences for Bengali-English from multiple sources, trained NMT models that outperformed previous results, and prepared a new test set; thus elevating Bengali from its low-resource status. In future, we plan to design segmentation-agnostic aligners or aligners that can jointly segment and align sentences. We want to experiment more with the LASER toolkit: we used LASER out-of-the-box, we want to train it with our data, and modify the model architecture to improve it further. LASER fails to identify one-to-many/many-to-one sentence alignments, we want to address this. We would also like to experiment with BERT \citep{devlin2018bert} embeddings for similarity search. Furthermore, we wish to explore semi-supervised and unsupervised approaches to leverage monolingual data and explore multilingual machine translation for low-resource Indic languages.

\section*{Acknowledgements}
We would like to thank the ICT Division, Government of the People's Republic of Bangladesh for funding the project and Intelligent Machines Limited for providing cloud support. 
\bibliography{anthology,emnlp2020}

\begin{thebibliography}{50}
\expandafter\ifx\csname natexlab\endcsname\relax\def\natexlab#1{#1}\fi

\bibitem[{Abdelali et~al.(2014)Abdelali, Guzman, Sajjad, and
  Vogel}]{abdelali2014amara}
Ahmed Abdelali, Francisco Guzman, Hassan Sajjad, and Stephan Vogel. 2014.
\newblock \href {https://www.aclweb.org/anthology/L14-1675} {The {AMARA}
  corpus: Building parallel language resources for the educational domain}.
\newblock In \emph{Proceedings of the Ninth International Conference on
  Language Resources and Evaluation (LREC 2014)}, pages 1856--1862, Reykjavik,
  Iceland. European Language Resources Association (ELRA).

\bibitem[{Abdul-Rauf et~al.(2012)Abdul-Rauf, Fishel, Lambert, Noubours, and
  Sennrich}]{abdul2012extrinsic}
Sadaf Abdul-Rauf, Mark Fishel, Patrik Lambert, Sandra Noubours, and Rico
  Sennrich. 2012.
\newblock \href {http://tm-town-nlp-resources.s3.amazonaws.com/sent-align.pdf}
  {Extrinsic evaluation of sentence alignment systems}.
\newblock In \emph{Proceedings of the Workshop on Creating Cross-language
  Resources for Disconnected Languages and Styles}, pages 6--10, Istanbul,
  Turkey.

\bibitem[{Agi{\'c} and Vuli{\'c}(2019)}]{agic2019jw300}
{\v{Z}}eljko Agi{\'c} and Ivan Vuli{\'c}. 2019.
\newblock \href {https://www.aclweb.org/anthology/P19-1310} {{JW}300: A
  wide-coverage parallel corpus for low-resource languages}.
\newblock In \emph{Proceedings of the 57th Annual Meeting of the Association
  for Computational Linguistics}, pages 3204--3210, Florence, Italy.
  Association for Computational Linguistics.

\bibitem[{Al-Rfou{'} et~al.(2013)Al-Rfou{'}, Perozzi, and
  Skiena}]{al2013polyglot}
Rami Al-Rfou{'}, Bryan Perozzi, and Steven Skiena. 2013.
\newblock \href {https://www.aclweb.org/anthology/W13-3520} {{P}olyglot:
  Distributed word representations for multilingual {NLP}}.
\newblock In \emph{Proceedings of the Seventeenth Conference on Computational
  Natural Language Learning}, pages 183--192, Sofia, Bulgaria. Association for
  Computational Linguistics.

\bibitem[{Artetxe and Schwenk(2019)}]{artetxe2018margin}
Mikel Artetxe and Holger Schwenk. 2019.
\newblock \href {https://www.aclweb.org/anthology/P19-1309} {Margin-based
  parallel corpus mining with multilingual sentence embeddings}.
\newblock In \emph{Proceedings of the 57th Annual Meeting of the Association
  for Computational Linguistics}, pages 3197--3203, Florence, Italy.
  Association for Computational Linguistics.

\bibitem[{Asaduzzaman and Ali(2003)}]{asaduzzaman2003morphological}
MM~Asaduzzaman and Muhammad~Masroor Ali. 2003.
\newblock Morphological analysis of {B}angla words for automatic machine
  translation.
\newblock In \emph{Proceedings of 6th International Conference on Computers and
  Information Technology (ICCIT)}, pages 271--276, Dhaka, Bangladesh.

\bibitem[{Bahdanau et~al.(2015)Bahdanau, Cho, and Bengio}]{bahdanau2014neural}
Dzmitry Bahdanau, Kyunghyun Cho, and Yoshua Bengio. 2015.
\newblock \href {http://arxiv.org/abs/1409.0473} {Neural machine translation by
  jointly learning to align and translate}.
\newblock In \emph{Proceedings of the 3rd International Conference on Learning
  Representations (ICLR 2015)}, San Diego, California, USA.

\bibitem[{Banerjee et~al.(2018)Banerjee, Kunchukuttan, and
  Bhattacharya}]{banerjee2018multilingual}
Tamali Banerjee, Anoop Kunchukuttan, and Pushpak Bhattacharya. 2018.
\newblock \href {https://www.aclweb.org/anthology/Y18-3013} {Multilingual
  {I}ndian language translation system at {WAT} 2018: Many-to-one phrase-based
  {SMT}}.
\newblock In \emph{Proceedings of the 32nd Pacific Asia Conference on Language,
  Information and Computation: 5th Workshop on Asian Translation}, Hong Kong.
  Association for Computational Linguistics.

\bibitem[{Braune and Fraser(2010)}]{braune2010improved}
Fabienne Braune and Alexander Fraser. 2010.
\newblock \href {https://www.aclweb.org/anthology/C10-2010} {Improved
  unsupervised sentence alignment for symmetrical and asymmetrical parallel
  corpora}.
\newblock In \emph{Proceedings of the 23rd International Conference on
  Computational Linguistics: Posters (COLING 2010)}, pages 81--89, Beijing,
  China. Association for Computational Linguistics.

\bibitem[{Cettolo et~al.(2012)Cettolo, Girardi, and Federico}]{cettolo2012wit3}
Mauro Cettolo, Christian Girardi, and Marcello Federico. 2012.
\newblock \href
  {https://cris.fbk.eu/retrieve/handle/11582/104409/4358/WIT3-EAMT2012.pdf}
  {Wit3: Web inventory of transcribed and translated talks}.
\newblock In \emph{Proceeding of the 16th Annual Conference of the European
  Association for Machine Translation (EAMT 2012)}, pages 261--268, Trento,
  Italy. European Association for Machine Translation.

\bibitem[{Chaudhary et~al.(2019)Chaudhary, Tang, Guzm{\'a}n, Schwenk, and
  Koehn}]{chaudhary2019low}
Vishrav Chaudhary, Yuqing Tang, Francisco Guzm{\'a}n, Holger Schwenk, and
  Philipp Koehn. 2019.
\newblock \href {https://www.aclweb.org/anthology/W19-5435} {Low-resource
  corpus filtering using multilingual sentence embeddings}.
\newblock In \emph{Proceedings of the Fourth Conference on Machine Translation
  (Volume 3: Shared Task Papers, Day 2)}, pages 261--266, Florence, Italy.
  Association for Computational Linguistics.

\bibitem[{Dandapat and Lewis(2018)}]{dandapat2018training}
Sandipan Dandapat and William Lewis. 2018.
\newblock \href {http://eamt2018.dlsi.ua.es/proceedings-eamt2018.pdf#page=129}
  {Training deployable general domain {MT} for a low resource language pair:
  {E}nglish--{B}angla}.
\newblock In \emph{Proceeding of the 21st Annual Conference of the European
  Association for Machine Translation (EAMT 2018)}, pages 109--117, Alacant,
  Spain. European Association for Machine Translation.

\bibitem[{Dasgupta et~al.(2004)Dasgupta, Wasif, and Azam}]{dasgupta2004optimal}
Sajib Dasgupta, Abu Wasif, and Sharmin Azam. 2004.
\newblock An optimal way of machine translation from {E}nglish to {B}engali.
\newblock In \emph{Proceedings of 7th International Conference on Computers and
  Information Technology (ICCIT)}, pages 648--653, Dhaka, Bangladesh.

\bibitem[{Devlin et~al.(2019)Devlin, Chang, Lee, and
  Toutanova}]{devlin2018bert}
Jacob Devlin, Ming-Wei Chang, Kenton Lee, and Kristina Toutanova. 2019.
\newblock \href {https://www.aclweb.org/anthology/N19-1423} {{BERT}:
  Pre-training of deep bidirectional transformers for language understanding}.
\newblock In \emph{Proceedings of the 2019 Conference of the North {A}merican
  Chapter of the Association for Computational Linguistics: Human Language
  Technologies, Volume 1 (Long and Short Papers)}, pages 4171--4186,
  Minneapolis, Minnesota, USA. Association for Computational Linguistics.

\bibitem[{Gu et~al.(2018)Gu, Hassan, Devlin, and Li}]{gu2018universal}
Jiatao Gu, Hany Hassan, Jacob Devlin, and Victor~O.K. Li. 2018.
\newblock \href {https://www.aclweb.org/anthology/N18-1032} {Universal neural
  machine translation for extremely low resource languages}.
\newblock In \emph{Proceedings of the 2018 Conference of the North {A}merican
  Chapter of the Association for Computational Linguistics: Human Language
  Technologies, Volume 1 (Long Papers)}, pages 344--354, New Orleans,
  Louisiana, USA. Association for Computational Linguistics.

\bibitem[{Guzm{\'a}n et~al.(2019)Guzm{\'a}n, Chen, Ott, Pino, Lample, Koehn,
  Chaudhary, and Ranzato}]{guzman2019two}
Francisco Guzm{\'a}n, Peng-Jen Chen, Myle Ott, Juan Pino, Guillaume Lample,
  Philipp Koehn, Vishrav Chaudhary, and Marc{'}Aurelio Ranzato. 2019.
\newblock \href {https://www.aclweb.org/anthology/D19-1632} {The {FLORES}
  evaluation datasets for low-resource machine translation:
  {N}epali{--}{E}nglish and {S}inhala{--}{E}nglish}.
\newblock In \emph{Proceedings of the 2019 Conference on Empirical Methods in
  Natural Language Processing and the 9th International Joint Conference on
  Natural Language Processing (EMNLP-IJCNLP)}, pages 6098--6111, Hong Kong,
  China. Association for Computational Linguistics.

\bibitem[{Haffari et~al.(2009)Haffari, Roy, and Sarkar}]{haffari2009active}
Gholamreza Haffari, Maxim Roy, and Anoop Sarkar. 2009.
\newblock \href {https://www.aclweb.org/anthology/N09-1047} {Active learning
  for statistical phrase-based machine translation}.
\newblock In \emph{Proceedings of Human Language Technologies: The 2009 Annual
  Conference of the North {A}merican Chapter of the Association for
  Computational Linguistics}, pages 415--423, Boulder, Colorado, USA.
  Association for Computational Linguistics.

\bibitem[{Hasan et~al.(2019)Hasan, Alam, Chowdhury, and Khan}]{hasanneural}
Md.~Arid Hasan, Firoj Alam, Shammur~Absar Chowdhury, and Naira Khan. 2019.
\newblock Neural machine translation for the {B}angla-{E}nglish language pair.
\newblock In \emph{Proceedings of 22nd International Conference on Computers
  and Information Technology (ICCIT)}, pages 1--6, Dhaka, Bangladesh.

\bibitem[{Irvine and Callison-Burch(2013)}]{irvine2013combining}
Ann Irvine and Chris Callison-Burch. 2013.
\newblock \href {https://www.aclweb.org/anthology/W13-2233} {Combining
  bilingual and comparable corpora for low resource machine translation}.
\newblock In \emph{Proceedings of the Eighth Workshop on Statistical Machine
  Translation}, pages 262--270, Sofia, Bulgaria. Association for Computational
  Linguistics.

\bibitem[{Islam et~al.(2010)Islam, Tiedemann, and Eisele}]{islam2010english}
Md~Zahurul Islam, J{\"o}rg Tiedemann, and Andreas Eisele. 2010.
\newblock \href {http://www.mt-archive.info/EAMT-2010-Islam.pdf} {{E}nglish to
  {B}angla phrase-based machine translation}.
\newblock In \emph{Proceeding of the 14th Annual Conference of the European
  Association for Machine Translation (EAMT 2010)}, St Raphael, France.
  European Association for Machine Translation.

\bibitem[{Johnson et~al.(2017)Johnson, Schuster, Le, Krikun, Wu, Chen, Thorat,
  Vi{\'e}gas, Wattenberg, Corrado, Hughes, and Dean}]{johnson2017google}
Melvin Johnson, Mike Schuster, Quoc~V. Le, Maxim Krikun, Yonghui Wu, Zhifeng
  Chen, Nikhil Thorat, Fernanda Vi{\'e}gas, Martin Wattenberg, Greg Corrado,
  Macduff Hughes, and Jeffrey Dean. 2017.
\newblock \href {https://www.aclweb.org/anthology/Q17-1024} {Google's
  multilingual neural machine translation system: Enabling zero-shot
  translation}.
\newblock \emph{Transactions of the Association for Computational Linguistics},
  5:339--351.

\bibitem[{Joulin et~al.(2017)Joulin, Grave, Bojanowski, and
  Mikolov}]{joulin2016bag}
Armand Joulin, Edouard Grave, Piotr Bojanowski, and Tomas Mikolov. 2017.
\newblock \href {https://www.aclweb.org/anthology/E17-2068} {Bag of tricks for
  efficient text classification}.
\newblock In \emph{Proceedings of the 15th Conference of the {E}uropean Chapter
  of the Association for Computational Linguistics: Volume 2, Short Papers},
  pages 427--431, Valencia, Spain. Association for Computational Linguistics.

\bibitem[{Khayrallah and Koehn(2018)}]{khayrallah2018impact}
Huda Khayrallah and Philipp Koehn. 2018.
\newblock \href {https://www.aclweb.org/anthology/W18-2709} {On the impact of
  various types of noise on neural machine translation}.
\newblock In \emph{Proceedings of the 2nd Workshop on Neural Machine
  Translation and Generation}, pages 74--83, Melbourne, Australia. Association
  for Computational Linguistics.

\bibitem[{Klein et~al.(2017)Klein, Kim, Deng, Senellart, and
  Rush}]{klein2017opennmt}
Guillaume Klein, Yoon Kim, Yuntian Deng, Jean Senellart, and Alexander Rush.
  2017.
\newblock \href {https://www.aclweb.org/anthology/P17-4012} {{O}pen{NMT}:
  Open-source toolkit for neural machine translation}.
\newblock In \emph{Proceedings of {ACL} 2017, System Demonstrations}, pages
  67--72, Vancouver, Canada. Association for Computational Linguistics.

\bibitem[{Koehn et~al.(2019)Koehn, Guzm{\'a}n, Chaudhary, and
  Pino}]{koehn2019findings}
Philipp Koehn, Francisco Guzm{\'a}n, Vishrav Chaudhary, and Juan Pino. 2019.
\newblock \href {https://www.aclweb.org/anthology/W19-5404} {Findings of the
  {WMT} 2019 shared task on parallel corpus filtering for low-resource
  conditions}.
\newblock In \emph{Proceedings of the Fourth Conference on Machine Translation
  (Volume 3: Shared Task Papers, Day 2)}, pages 54--72, Florence, Italy.
  Association for Computational Linguistics.

\bibitem[{Koehn et~al.(2007)Koehn, Hoang, Birch, Callison-Burch, Federico,
  Bertoldi, Cowan, Shen, Moran, Zens, Dyer, Bojar, Constantin, and
  Herbst}]{koehn2007moses}
Philipp Koehn, Hieu Hoang, Alexandra Birch, Chris Callison-Burch, Marcello
  Federico, Nicola Bertoldi, Brooke Cowan, Wade Shen, Christine Moran, Richard
  Zens, Chris Dyer, Ond{\v{r}}ej Bojar, Alexandra Constantin, and Evan Herbst.
  2007.
\newblock \href {https://www.aclweb.org/anthology/P07-2045} {{M}oses: Open
  source toolkit for statistical machine translation}.
\newblock In \emph{Proceedings of the 45th Annual Meeting of the Association
  for Computational Linguistics Companion Volume Proceedings of the Demo and
  Poster Sessions}, pages 177--180, Prague, Czech Republic. Association for
  Computational Linguistics.

\bibitem[{Koehn and Knowles(2017)}]{koehn2017six}
Philipp Koehn and Rebecca Knowles. 2017.
\newblock \href {https://www.aclweb.org/anthology/W17-3204} {Six challenges for
  neural machine translation}.
\newblock In \emph{Proceedings of the First Workshop on Neural Machine
  Translation}, pages 28--39, Vancouver, Canada. Association for Computational
  Linguistics.

\bibitem[{Koehn et~al.(2003)Koehn, Och, and Marcu}]{koehn2003statistical}
Philipp Koehn, Franz~J. Och, and Daniel Marcu. 2003.
\newblock \href {https://www.aclweb.org/anthology/N03-1017} {Statistical
  phrase-based translation}.
\newblock In \emph{Proceedings of the 2003 Human Language Technology Conference
  of the North {A}merican Chapter of the Association for Computational
  Linguistics}, pages 127--133.

\bibitem[{Kudo(2018)}]{kudo2018subword}
Taku Kudo. 2018.
\newblock \href {https://www.aclweb.org/anthology/P18-1007} {Subword
  regularization: Improving neural network translation models with multiple
  subword candidates}.
\newblock In \emph{Proceedings of the 56th Annual Meeting of the Association
  for Computational Linguistics (Volume 1: Long Papers)}, pages 66--75,
  Melbourne, Australia. Association for Computational Linguistics.

\bibitem[{Kudo and Richardson(2018)}]{kudo2018sentencepiece}
Taku Kudo and John Richardson. 2018.
\newblock \href {https://www.aclweb.org/anthology/D18-2012} {{S}entence{P}iece:
  A simple and language independent subword tokenizer and detokenizer for
  neural text processing}.
\newblock In \emph{Proceedings of the 2018 Conference on Empirical Methods in
  Natural Language Processing: System Demonstrations}, pages 66--71, Brussels,
  Belgium. Association for Computational Linguistics.

\bibitem[{Lamraoui and Langlais(2013)}]{lamraoui2013yet}
Fethi Lamraoui and Philippe Langlais. 2013.
\newblock \href {http://www.mt-archive.info/10/MTS-2013-Lamraoui.pdf} {Yet
  another fast, robust and open source sentence aligner. time to reconsider
  sentence alignment}.
\newblock In \emph{Proceedings of the XIV Machine Translation Summit}, pages
  77--84, Nice, France.

\bibitem[{Lison et~al.(2018)Lison, Tiedemann, and Kouylekov}]{lison2019open}
Pierre Lison, J{\"o}rg Tiedemann, and Milen Kouylekov. 2018.
\newblock \href {https://www.aclweb.org/anthology/L18-1275}
  {{O}pen{S}ubtitles2018: Statistical rescoring of sentence alignments in
  large, noisy parallel corpora}.
\newblock In \emph{Proceedings of the Eleventh International Conference on
  Language Resources and Evaluation (LREC 2018)}, pages 1742--1748, Miyazaki,
  Japan. European Language Resources Association (ELRA).

\bibitem[{Mumin et~al.(2018)Mumin, Seddiqui, Iqbal, and Islam}]{czes-gs42-18}
Md~Abdullah~Al Mumin, Md~Hanif Seddiqui, Muhammed~Zafar Iqbal, and
  Mohammed~Jahirul Islam. 2018.
\newblock \href {https://doi.org/10.21227/czes-gs42} {{SUPara}-benchmark: A
  benchmark dataset for {E}nglish-{B}angla machine translation}.
\newblock In \emph{IEEE Dataport}.

\bibitem[{Mumin et~al.(2019{\natexlab{a}})Mumin, Seddiqui, Iqbal, and
  Islam}]{mumin2019neural}
Md~Abdullah~Al Mumin, Md~Hanif Seddiqui, Muhammed~Zafar Iqbal, and
  Mohammed~Jahirul Islam. 2019{\natexlab{a}}.
\newblock \href {https://doi.org/10.3844/jcssp.2019.1627.1637} {Neural machine
  translation for low-resource {E}nglish-{B}angla}.
\newblock \emph{Journal of Computer Science}, 15(11):1627--1637.

\bibitem[{Mumin et~al.(2019{\natexlab{b}})Mumin, Seddiqui, Iqbal, and
  Islam}]{mumin2019shutorjoma}
Md~Abdullah~Al Mumin, Md~Hanif Seddiqui, Muhammed~Zafar Iqbal, and
  Mohammed~Jahirul Islam. 2019{\natexlab{b}}.
\newblock \href {https://doi.org/10.3844/jcssp.2019.1022.1039} {shu-torjoma: An
  {E}nglish $\leftrightarrow$ {B}angla statistical machine translation system}.
\newblock \emph{Journal of Computer Science}, 15(7):1022--1039.

\bibitem[{Mumin et~al.(2012)Mumin, Shoeb, Selim, and Iqbal}]{al2012supara}
Md~Abdullah~Al Mumin, Abu Awal~Md Shoeb, Md~Reza Selim, and Muhammed~Zafar
  Iqbal. 2012.
\newblock \href
  {http://journals.sust.edu/uploads/archive/8db409b8441d7fddff7755c9859372c3.pdf}
  {{SUPara}: a balanced {E}nglish-{B}engali parallel corpus}.
\newblock \emph{SUST Journal of Science and Technology}, 16(2):46--51.

\bibitem[{Naskar and Bandyopadhyay(2005)}]{naskar2006phrasal}
Sudip~Kumar Naskar and Sivaji Bandyopadhyay. 2005.
\newblock \href {http://www.mt-archive.info/MTS-2005-Naskar-1.pdf} {A phrasal
  {EBMT} system for translating {E}nglish to {B}engali}.
\newblock In \emph{Proceedings of the Tenth Machine Translation Summit}, pages
  372--279, Phuket, Thailand.

\bibitem[{Papineni et~al.(2002)Papineni, Roukos, Ward, and
  Zhu}]{papineni2002bleu}
Kishore Papineni, Salim Roukos, Todd Ward, and Wei-Jing Zhu. 2002.
\newblock \href {https://www.aclweb.org/anthology/P02-1040} {{B}leu: a method
  for automatic evaluation of machine translation}.
\newblock In \emph{Proceedings of the 40th Annual Meeting of the Association
  for Computational Linguistics}, pages 311--318, Philadelphia, Pennsylvania,
  USA. Association for Computational Linguistics.

\bibitem[{Post(2018)}]{post2018call}
Matt Post. 2018.
\newblock \href {https://www.aclweb.org/anthology/W18-6319} {A call for clarity
  in reporting {BLEU} scores}.
\newblock In \emph{Proceedings of the Third Conference on Machine Translation:
  Research Papers}, pages 186--191, Brussels, Belgium. Association for
  Computational Linguistics.

\bibitem[{Post et~al.(2012)Post, Callison-Burch, and
  Osborne}]{post2012constructing}
Matt Post, Chris Callison-Burch, and Miles Osborne. 2012.
\newblock \href {https://www.aclweb.org/anthology/W12-3152} {Constructing
  parallel corpora for six {I}ndian languages via crowdsourcing}.
\newblock In \emph{Proceedings of the Seventh Workshop on Statistical Machine
  Translation}, pages 401--409, Montr{\'e}al, Canada. Association for
  Computational Linguistics.

\bibitem[{Roy(2009)}]{roy2009semi}
Maxim Roy. 2009.
\newblock \href {https://doi.org/10.1007/978-3-642-01818-3_45} {A
  semi-supervised approach to {B}engali-{E}nglish phrase-based statistical
  machine translation}.
\newblock In \emph{Proceedings of the 22nd Canadian Conference on Artificial
  Intelligence: Advances in Artificial Intelligence}, page 291–294, Kelowna,
  Canada. Springer-Verlag.

\bibitem[{Saha and Bandyopadhyay(2005)}]{saha2005semantics}
Diganta Saha and Sivaji Bandyopadhyay. 2005.
\newblock \href {http://www.mt-archive.info/MTS-2005-Saha.pdf} {A
  semantics-based {E}nglish-{B}engali {EBMT} system for translating news
  headlines}.
\newblock In \emph{Proceedings of the Tenth Machine Translation Summit}, pages
  125--133, Phuket, Thailand.

\bibitem[{Schwenk et~al.(2019)Schwenk, Chaudhary, Sun, Gong, and
  Guzm{\'a}n}]{schwenk2019wikimatrix}
Holger Schwenk, Vishrav Chaudhary, Shuo Sun, Hongyu Gong, and Francisco
  Guzm{\'a}n. 2019.
\newblock \href {https://arxiv.org/abs/1907.05791} {Wikimatrix: Mining 135m
  parallel sentences in 1620 language pairs from wikipedia}.
\newblock \emph{arXiv:1907.05791}.

\bibitem[{Sennrich et~al.(2016)Sennrich, Haddow, and
  Birch}]{sennrich2015improving}
Rico Sennrich, Barry Haddow, and Alexandra Birch. 2016.
\newblock \href {https://www.aclweb.org/anthology/P16-1009} {Improving neural
  machine translation models with monolingual data}.
\newblock In \emph{Proceedings of the 54th Annual Meeting of the Association
  for Computational Linguistics (Volume 1: Long Papers)}, pages 86--96, Berlin,
  Germany. Association for Computational Linguistics.

\bibitem[{Sennrich and Volk(2010)}]{sennrich2010mt}
Rico Sennrich and Martin Volk. 2010.
\newblock \href {http://www.mt-archive.info/10/AMTA-2010-Sennrich.pdf}
  {{MT}-based sentence alignment for {OCR}-generated parallel texts}.
\newblock In \emph{Proceedings of The Ninth Conference of the Association for
  Machine Translation in the Americas (AMTA 2010)}, Denver, Colorado, USA.

\bibitem[{Sinha et~al.(1995)Sinha, Sivaraman, Agrawal, Jain, Srivastava, and
  Jain}]{sinha1995anglabharti}
RMK Sinha, K~Sivaraman, Aditi Agrawal, Renu Jain, Rakesh Srivastava, and Ajai
  Jain. 1995.
\newblock \href {https://ieeexplore.ieee.org/document/538002} {{ANGLABHARTI}: a
  multilingual machine aided translation project on translation from {E}nglish
  to {I}ndian languages}.
\newblock In \emph{1995 IEEE International Conference on Systems, Man and
  Cybernetics. Intelligent Systems for the 21st Century}, volume~2, pages
  1609--1614.

\bibitem[{Tiedemann(2012)}]{tiedemann2012parallel}
J{\"o}rg Tiedemann. 2012.
\newblock \href {https://www.aclweb.org/anthology/L12-1246} {Parallel data,
  tools and interfaces in {OPUS}}.
\newblock In \emph{Proceedings of the Eighth International Conference on
  Language Resources and Evaluation (LREC 2012)}, pages 2214--2218, Istanbul,
  Turkey. European Language Resources Association (ELRA).

\bibitem[{Varga et~al.(2005)Varga, Hal{\'a}csy, Kornai, Nagy, N{\'e}meth, and
  Tr{\'o}n}]{varga2007parallel}
D{\'a}niel Varga, P{\'e}ter Hal{\'a}csy, Andr{\'a}s Kornai, Viktor Nagy,
  L{\'a}szl{\'o} N{\'e}meth, and Viktor Tr{\'o}n. 2005.
\newblock \href {https://catalog.ldc.upenn.edu/docs/LDC2008T01/ranlp05.pdf}
  {Parallel corpora for medium density languages}.
\newblock In \emph{Proceedings of the Recent Advances in Natural Language
  Processing (RANLP 2005)}, Borovets, Bulgaria.

\bibitem[{Vaswani et~al.(2017)Vaswani, Shazeer, Parmar, Uszkoreit, Jones,
  Gomez, Kaiser, and Polosukhin}]{vaswani2017attention}
Ashish Vaswani, Noam Shazeer, Niki Parmar, Jakob Uszkoreit, Llion Jones,
  Aidan~N Gomez, {\L}ukasz Kaiser, and Illia Polosukhin. 2017.
\newblock \href
  {https://papers.nips.cc/paper/7181-attention-is-all-you-need.pdf} {Attention
  is all you need}.
\newblock In \emph{Proceedings of the 31st International Conference on Neural
  Information Processing Systems (NIPS 2017)}, page 6000–6010, Long Beach,
  California, USA.

\bibitem[{Wu et~al.(2016)Wu, Schuster, Chen, Le, Norouzi, Macherey, Krikun,
  Cao, Gao, Macherey et~al.}]{wu2016google}
Yonghui Wu, Mike Schuster, Zhifeng Chen, Quoc~V Le, Mohammad Norouzi, Wolfgang
  Macherey, Maxim Krikun, Yuan Cao, Qin Gao, Klaus Macherey, et~al. 2016.
\newblock \href {https://arxiv.org/abs/1609.08144} {Google's neural machine
  translation system: Bridging the gap between human and machine translation}.
\newblock \emph{arXiv:1609.08144}.

\end{thebibliography}
\bibliographystyle{acl_natbib}
\end{document}